\documentclass{article}
\usepackage{ijcai20}

\usepackage{times}
\usepackage{soul}
\usepackage{url}
\usepackage{adjustbox}
\usepackage[hidelinks]{hyperref}
\usepackage[utf8]{inputenc}
\usepackage[table]{xcolor}
\usepackage{makecell}
\usepackage[small]{caption}
\usepackage{graphicx}
\usepackage{amsmath}
\usepackage{amsthm}
\usepackage{subfig}
\usepackage{booktabs}
\usepackage{algorithm}
\usepackage{algorithmic}
\title{Single Run Action Detector over Video Stream - A Privacy Preserving Approach}

\author{
Anbumalar Saravanan$^1$
\and
Justin Sanchez$^1$\and
Hassan Ghasemzadeh$^2$\and
Aurelia Macabasco-O'Connell$^3$\And
Hamed Tabkhi$^{1}$
\affiliations
$^1$University of North Carolina Charlotte\\
$^2$University of North Carolina Charlotte\\
$^3$University of North Carolina Charlotte\\
$^2$School of EECS Washington State University\\
$^3$School of Nursing Azusa Pacific University\\
$^4$School of Nursing Azusa Pacific University\\
\emails
\{asaravan, jsanch19,htabkhiv\}@uncc.edu,
hassan.ghasemzadeh@wsu.edu
\{amacabascooconnell\}@apu.edu
}

\begin{document}

\maketitle

\begin{abstract}
  This paper takes initial strides at designing and evaluating a vision-based system for privacy ensured activity monitoring. The proposed technology utilizing Artificial Intelligence (AI)-empowered proactive systems offering continuous monitoring, behavioral analysis, and modeling of human activities. To this end, this paper presents Single Run Action Detector (S-RAD) which is a real-time privacy-preserving action detector that performs end-to-end action localization and classification. It is based on Faster-RCNN combined with temporal shift modeling and segment based sampling to capture the human actions. Results on UCF-Sports and UR Fall dataset present comparable accuracy to State-of-the-Art approaches with significantly lower model size and computation demand and the ability for real-time execution on edge embedded device (e.g. Nvidia Jetson Xavier).
\end{abstract}

\section{Introduction}

In recent years, deep learning has achieved success in fields such as computer vision and natural language processing. Compared to traditional machine learning methods such as support vector and random forest, deep learning has a strong learning ability from the data and can make better use of datasets for feature extraction. Because of this practicability, deep learning had become more and more popular to do research works.

Deep learning models usually adopt hierarchical structures to connect their layers. The output of a lower layer can be regarded as the input of a higher layer using linear or nonlinear functions. These models can transform low-level features to high-level abstract features from the input data. Because of this characteristic, deep learning models are stronger than shallow machine learning models in feature representation. The performance of traditional machine-learning methods usually rely on user experiences and handcrafted methods, while deep learning approaches rely on the data. 

The recent approaches in video analytic and deep learning algorithms like Convolutional Neural network provides the opportunity for real-time detection and analysis of human behaviors like walking,running or sitting down, which are part of daily living Activities (ADL) \cite{REVAMP2T}. Cameras provide very rich information about persons and environments and their presence is becoming more important in everyday environments like airports, train and bus stations,malls,elderly care and even streets. Therefore, reliable vision-based action detection systems is required for various application like healthcare assistance system, crime detection and sports monitoring system. In our paper we explored two different domains(Sport and Healthcare), to prove the comprehensive nature of our proposed action detector algorithm. Approaches like \cite{Cameiro,Alaouikeypointfalldetection,videocapsule_net,TCNN} use larger CNN models that impose huge computation demand and thus limit their application in real-time constrained systems, in particular on embedded edge devices. Additionally,these methods have not been designed to fulfill requirements of pervasive video systems including privacy-preserving and real-time responsiveness. Other works done in this area are based on the use of wearable sensors. These works used the tri-axial accelerometer, ambient/fusion, vibrations or audio and video to capture the human posture,body shape change. However, wearable sensors require relative strict positioning and thus bring along inconvenience especially in the scenario of healthcare unit where elderly seniors may even forget to wear them.

Motivated by the need and importance of image based action detection system we introduce a novel Single Run Action detector (S-RAD) for activity monitoring. S-RAD provides end-to-end action detection without the use of computationally heavy methods in a single shot manner with the ability to run real-time on embedded edge device. S-RAD detects and localizes complex human actions with a Faster-RCNN like architecture \cite{Faster-RCNN} combined with temporal shift blocks (based on \cite{tsm}) to capture the low-level and high-level video temporal context. S-RAD is a privacy-preserving approach and inherently protects Personally Identifiable Information (PII). The real-time execution on edge avoids unnecessary video transfer and PII to a cloud or remote computing server.

Overall, our contributions are as follows:
(1) We introduce S-RAD, a single shot action detector localising humans and classifying actions. (2) We demonstrate that we can achieve comparable accuracy to the State-of-the-Art approaches (on the UCF-Sports and UR Fall datasets) at much lower computation cost. We demonstrate our approach on two different dataset from healthcare and sport domain to prove it's robustness and applicability to multiple action detection domains. (3) We additionally provide possibility's of extending our network to real-time scenarios on an edge device. Code will be made publicly available on GitHub after reviews.

\footnote{This is a pre-print of an article to be published in the 2nd International Workshop On Deep Learning For Human Activity Recognition, Springer Communications in Computer and Information Science (CCIS) proceedings.}

\section{Related Works}

\subsection{Activity Recognition using Wearable Sensors:} Most prior research focuses on using wearable and mobile devices (e.g., smartphones, smartwatches) for activity recognition . In designing efficient activity recognition systems, researchers have extensively studied various wearable computing research questions. These research efforts have revolved around optimal placement of the wearable sensors \cite{atallah2011sensor}, automatic detection of the on-body location of the sensor \cite{saeedi2014toward}, minimization of the sensing energy consumption \cite{pagan2018toward}, and optimization of the power consumption \cite{Mirzadeh2020OptDeploy}. A limitation of activity monitoring using wearable sensors and mobile devices is that these technologies are battery-powered and therefore need to be regularly charged. Failure to charge the battery results in discontinuity of the activity recognition, which in turn may lead to important behavioral events remaining undetected. 
\subsection{Action Recognition in Video data}
Action recognition is a long-term research problem and has been studied for decades. Existing State-of-the-Art methods mostly focus on modelling the temporal dependencies in the successive video frames \cite{Twostream,TSN,c3D}. For instance, \cite{TSN} directly averaged the motion cues depicted in different temporal segments in order to capture the irregular nature of temporal information. \cite{Twostream} proposed a two-stream network, which takes RGB frames and optical flows as input respectively and fused the detection's from the two streams as the final output. This was done at several granularities of abstraction and achieved great performance. Beyond multi-stream based methods, methods like \cite{c3D,3Dfalldetection} explored 3D ConvNets on video streams for joint spatio-temporal feature learning on videos. In this way, they avoid calculating the optical flow, keypoints or saliency maps explicitly. However all the above approaches are too large to fit in a real-time edge device.
On the other hand \cite{Alaouikeypointfalldetection} uses features calculated from variations in the human keypoints to classify falling and not falling actions, \cite{Cameiro} uses VGG16 based on Multi-stream (optical flow, RGB, pose estimation) for human action classification. The above approaches only concentrate on the classification of single human action at scene level and will not perform well if multiple human's are present in an image, which is essential for the healthcare and other public place monitoring systems. Our proposed approach performs human detection and action classification together in a single shot manner where algorithm first localises the human's in an image and classifies his/her action.
\subsection{Spatio-temporal Human Action Detection:}
Spatio-temporal human action detection is a challenging computer vision problem, which involves detecting human actions in a video as well as localizing these actions both spatially and temporally. Few papers on spatio-temporal action detection like \cite{ACTdetector} uses object detectors like SSD \cite{SSD} to generate spatio-temporal tubes by deploying high level linking algorithm on frame level detection's. Inspired by RCNN approaches, \cite{Multiregion} used Faster-RCNN \cite{Faster-RCNN} to detect the human in an image by capturing the action motion cues with the help of optical flow and classify the final human actions based on the actionness score. \cite{action_tubes} extracted proposals by using the selective search method on RGB frames and then applied the original R-CNN on per frame RGB and optical flow data for frame-level action detection's and finally link those detection's using the Viterbi algorithm to generate action tubes. 
On the other hand \cite{TCNN} uses 3D CNN to generate spatio-temporal tubes with Tube of interest pooling and had showed good performance in the action related datasets. However all these methods poses high processing time and computation cost due to optical flow generation in the two stream networks, 3D kernels in the 3D CNN related works and generation of keypoint's in the human pose based methods. As such, the aforementioned methods are unable to be applied in real-time monitoring systems.

\section{Single Run-Action Detector}
\textbf{Approach}
We introduce S-RAD, an agile and real-time activity monitoring system. Our approach unifies spatio-temporal feature extraction and localization into a single network, allowing the opportunity to be deployed on edge device. This "on-the-edge" deployment eliminates the need for sending sensitive human data to privacy invalidating cloud servers, similar to \cite{REVAMP2T}. Instead our approach can delete all video data after it is processed and can store only the high level activity analytics. Without stored images, S-RAD can be used to solely focus on differentiating between the human actions rather than identifying or describing the human.

\begin{figure*}[http]
    \centering
    \includegraphics[width=1\linewidth]{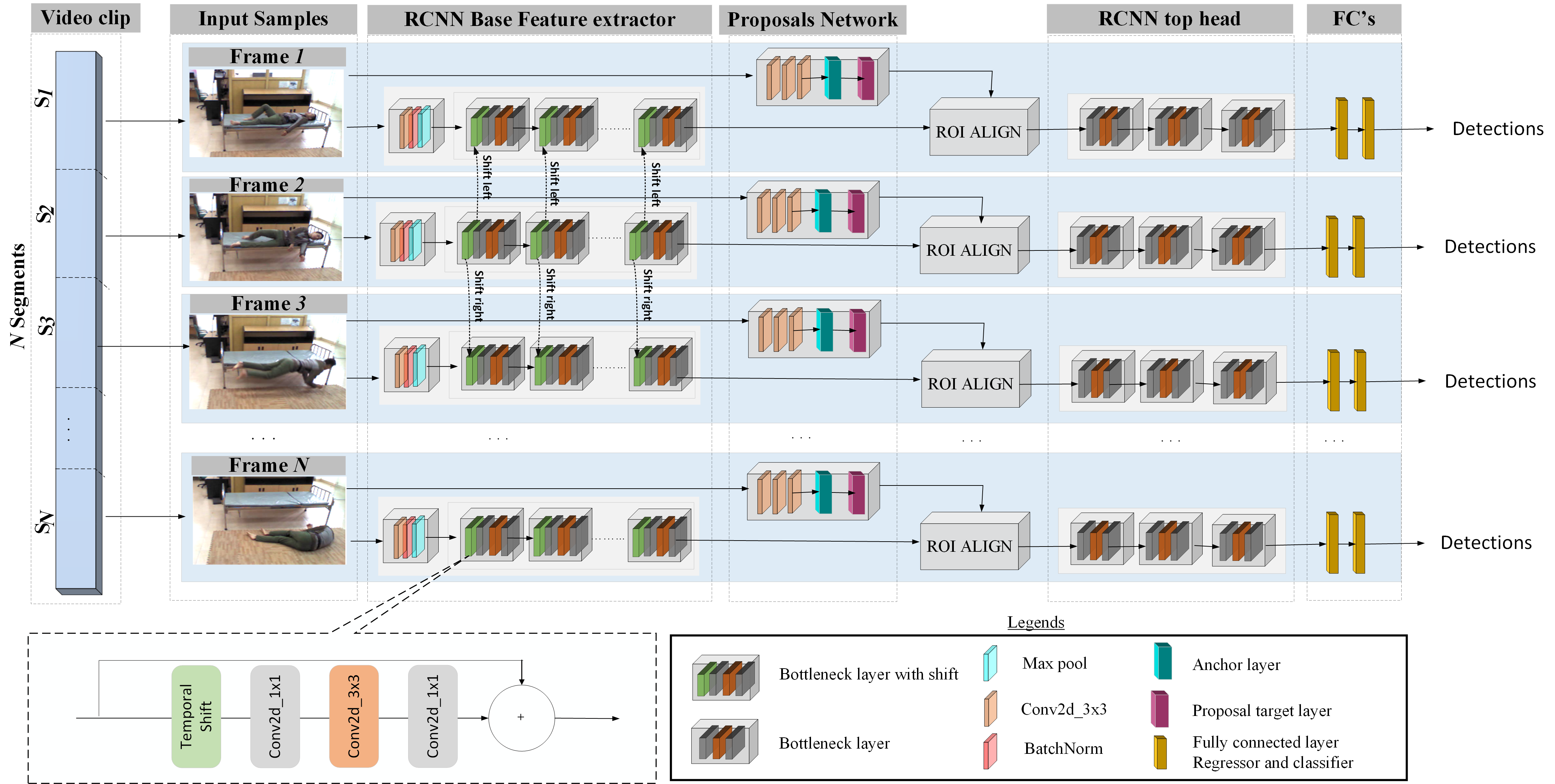}
    \caption{Overview of the activity detector. Given a sequence of frames we extract channel shifted convolutional features from the base feature extractor to derive the \emph{activity proposals} in the action proposal network. We then ROI align the activity proposals to predict their scores and regress their co-ordinates. }
        \vspace{-10pt}
    \label{fig:Action detector}
\end{figure*}

In order to achieve this privacy preserving edge execution, it is important to have an algorithm able to perform in a resource constrained edge environment. Traditionally such constraints resulted in either accuracy reduction, or increased latency.
The overview of S-RAD is shown in Figure {\ref{fig:Action detector}}. S-RAD takes an input sequence of N frames $f_1,f_2,f_3,...,f_N$ and outputs the detected bounding box and confidence score per each class of the proposals. The model consists of a base feature extractor integrated with temporal shift blocks to capture low level spatio-temporal features. The base feature extractor is made up of the first 40 layers of the original ResNet-50 \cite{resnet} backbone. The base feature maps are processed by the Region Proposal Network (RPN) using a sliding window approach with handpicked anchors and generates action proposals for each frame. An RPN is a fully convolutional network that simultaneously predicts action bounds and actionness scores at each position. The RPN is trained end-to-end to localize and detect valid region action proposals (the foreground) from background. This sliding window approach to generate the proposals is the source of its accuracy as opposed to SSD's \cite{SSD} rigid grid base proposal generation.
\begin{figure}[h!]
    \centering
     \includegraphics[width=1\linewidth]{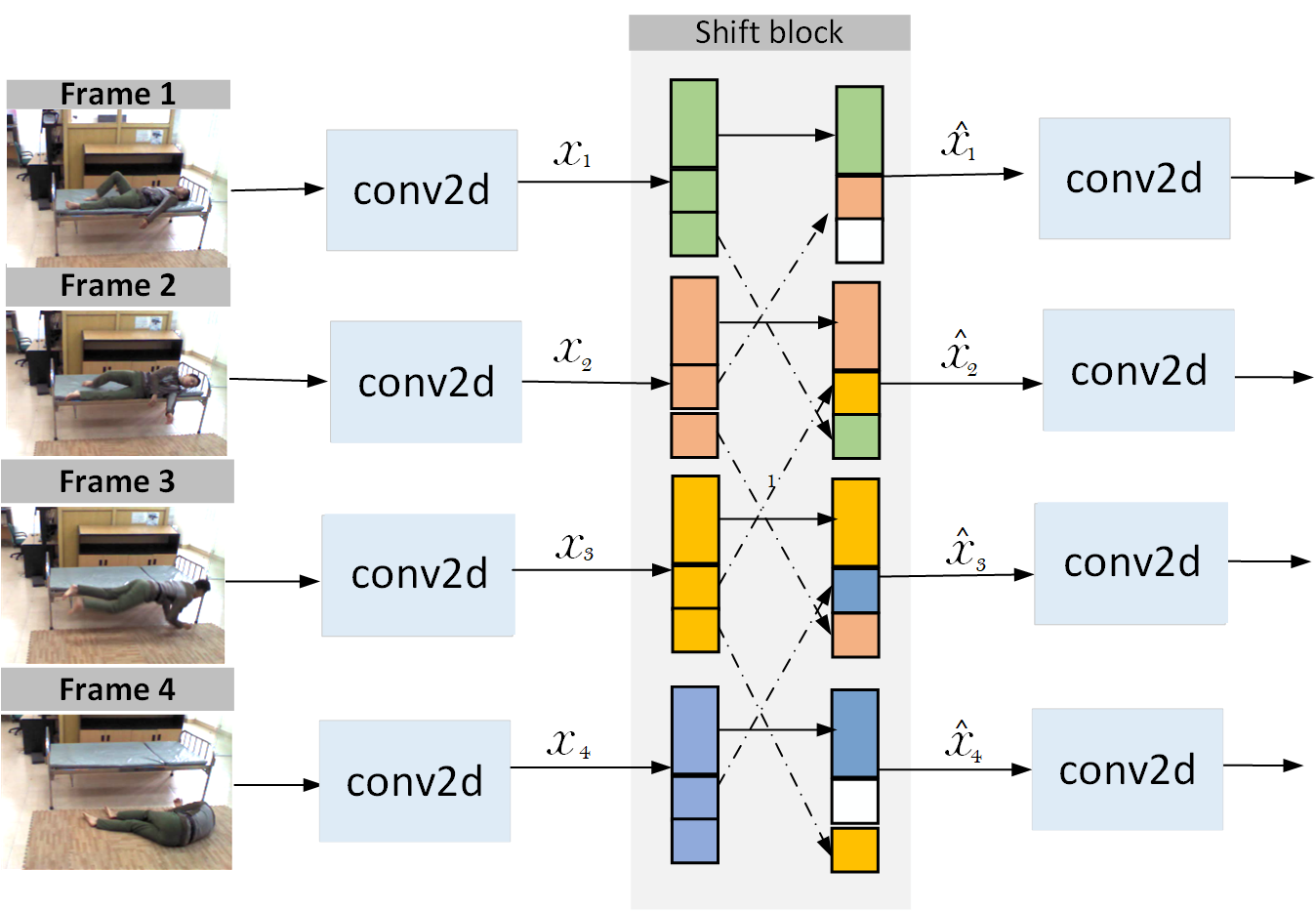}
    \caption{Temporal shift block}
    \label{fig:TemporalFeature}
\end{figure}
Following the first stage, the original spatio-temporal base features, in conjecture with the proposals are passed into the Region of interest Align (ROI-Align) layer which aligns the varying sized action proposals in to a fixed 7x7 spatial sized action proposals. The second stage of the action detector further classifies each valid action proposals to the action classes in that particular frame. The final classification layer outputs \emph{C}+1 scores for each action proposal, one per each action class plus one for the background. The regression layer outputs 4 x \emph{K} where \emph{K} is the number of action proposals generated in each frame.

\textbf{Temporal shift block} TSM \cite{tsm} are highly hardware efficient. Temporal shift blocks are inserted into the bottleneck layer of Resnet-50 \cite{resnet} based feature extractor to sustain the spatial information using the identity mapping along with the temporal information using the shifted features. As shown in Figure \ref{fig:TemporalFeature}, each shift receives the $C$ channels from the previous layer. We shift 1/8th of the channels from the past frame to the current frame and shift 1/8th of the channels from current frame to the future frame, while the other part of the channels remain unshifted. The new features (channels are referred to as features) $\hat{x}_2$, have the information of both the past $x_1$ and future $x_2$ frames after the "shift" operation. The features are convoluted and mixed into new spatio-temporal features. The shift block coupled to the next layer will do the same operation. Each shift block increases the temporal receptive field by a magnitude of 2 neighbor frames until $N$ frames. For our work we choose $N=8$ since features are in the magnitude of 8 in Resnet-50 architecture \cite{resnet}.

S-RAD goes beyond action classification to action detection. This is valuable for communal areas such as mesh halls, and for interactions with other human's and with objects. We chose Faster-RCNN \cite{Faster-RCNN} as our detection baseline due to its fine-grained detection capabilities when compared to SSD \cite{SSD}. This fine grained detection is especially applicable to the healthcare domain when dealing with wandering patients and fine-grain abnormal behaviors. Despite the complexity of such tasks our utilization of TSM \cite{tsm} enables the extraction of the necessary spatio-temporal features for human action localization and individual action classification, in a streaming real-time manner while maintaining privacy. 
\subsection{Training Loss}
\textbf{RPN Loss:} 
For training RPNs, we assign a binary action class label (of being an action or not i.e foreground vs background) to each anchor. We assign a positive action class label to two kinds of anchors:(i) the anchors with the highest Intersection-over Union (IoU) overlap with a ground-truth box, or (ii) an anchor that has an IoU $>$ 0.7 with any ground-truth box. We assign a negative action class label to a non-positive anchor if it's IoU $<$ 0.3 for all ground-truth boxes. Anchors that are neither positive nor negative do not contribute to the training. With these definitions, our loss function for RPN is defined as:
\begin{equation}
\small
\resizebox{1\hsize}{!}{
$L_{rpn}(\{p_i\},\{bb_i\}) =\frac{1}{K}.\sum\limits_{i=1}^{K} L_{cls}(p_i,p_i^*)+\frac{1}{K}.\sum\limits_{i=1}^{K}p_i^*L_{reg}(bb_i,bb_i^*)$
}
\label{eq:1}
\end{equation}
Here, $i$ is the index of an anchor in a mini-batch and $p_i$ is the predicted probability of anchor $i$ belonging to an action class. The ground-truth label $p^*_i$ is 1 if the anchor is positive, and 0 if the anchor is negative. The vector representing the 4 coordinates of the predicted bounding box is $bb_i$, and $bb^*_i$ is the ground-truth box associated with a positive anchor. The term $p^*_i$ $L_{reg}$ dictates the smooth L1 regression loss 
is activated only for positive anchors ($p^*_i = 1$) and is disabled otherwise ($p^*_i = 0$). $L_{cls}$ is log loss(cross-entropy) over two classes (action vs. no  action) and is averaged over K frames.
\newline
\textbf{RCNN loss:} 
The seconds stage of the detector assigns the action class label to the region of interest or foreground proposals from the RPN training. It involves classification loss and regression loss. The classification layer here includes detecting the correct action class label for the proposals from ROI align layer and regression layer is to regress the detected box with ground truth. The RCNN loss is defined as :
\begin{equation}
\small
\resizebox{1\hsize}{!}{
$L_{rcnn}(\{p_i\},\{bb_i\}) = \frac{1}{K}. \sum\limits_{i=1}^{K} L_{cls}(p_i,p_i^*)+\frac{1}{K}.\sum\limits_{i=1}^{K} L_{reg}(bb_i,bb_i^*)$\label{eq:1}
}
\end{equation}
where $i$ is the index of proposals or region of interests with spatial dimension $7$x$7$ and $p_i$ is the predicted probability of the action class label, with $p^*_i$ being the ground truth class label. The vector representing the 4 coordinates of the predicted bounding box is $bb_i$, and $bb^*_i$ is that of the ground-truth box. $L_{cls}$ is log loss (cross-entropy) over multi-classes, $L_{reg}$ is the smooth L1 regression loss and is averaged over K frames. In training mode we set the network to output 256 proposals and in inference mode network outputs 300 proposals.

\textbf{Total training loss:} Total loss is defined as sum of RCNN and RPN loss:
\begin{equation}
\small
\resizebox{0.9\hsize}{!}{
$Total\_loss=L_{rpn}(\{p_i\},\{bb_i\})+L_{rcnn}(\{p_i\},\{bb_i\})$
}
\end{equation}

\section{Results and Evaluations}

\textbf{Setup} We use Resnet-50 \cite{resnet} as the backbone of our architecture because of the network depth and residual connections that enable feature reuse and propagation. The UCF-Sports \cite{UCFsports} and UR Fall \cite{urfall} datasets are too small and are prone to over fitting, so we fine-tuned our network from Kinetics \cite{kay2017kinetics} pre-trained weights and froze the batch normalization layers. The training parameters for the UCF-Sports \cite{UCFsports} dataset are 300 training epochs, with an inital learning rate of 0.03 and a weight decay 0.1 every 60 epochs.  We utilized gradient accumulation with a batch size of 4 and an accumulation step of 3 to fit a total batch of 12 on one V100GPU. The training parameters for the UR Fall dataset \cite{urfall} are 80 training epochs, with initial learning rate of 0.02 and a weight decay 0.1 every 20 epochs. We use the uniform temporal sampling strategy done in \cite{TSN} to sample 8 frames from the video and resize the input resolution of the image to 300x400 for State-of-the-Art comparison. We used datasets from two different domain (Sport and Healthcare) to show the generic capability of our algorithm.

\subsection{Results on UCF-Sports Dataset}
The UCF-Sports dataset \cite{UCFsports} consists of 150 videos from 10 action classes. All videos have spatio-temporal annotations in the form of frame-level bounding boxes and we follow the same training/testing split used by \cite{action_tubes}. On average there are 103 videos in the training dataset and 47 videos in the testing dataset. Videos are truncated to the action and bounding boxes annotations are provided for all frames.
To quantify our results, we report the mean Average Precision (mAP) at the frame level (frame mAP). Frame-level metrics allow us to compare the quality of the detection's independently. We use the Precision-recall AUC (Area under curve) to calculate the average precision per class. We compute the mean of the average precision per class to see how much our algorithm is able to differentiate the features between action classes. We followed the same procedure as in the PASCAL VOC detection challenge \cite{pascal_voc} to have an apple to apple comparison with the State-of-the-Art approaches in the detection task.
We first evaluate S-RAD on the widely used UCF-Sports dataset. Table \ref{tab:Table1} indicates frame level Average Precision per class for an intersection-over-union threshold of 0.5. Our approach achieves a mean AP of 85.04\% . While obtaining excellent performance on most of the classes, walking is the only action for which the framework fails to detect the humans (40.71\% frame-AP). This is possibly due to several factors, the first being that the test videos for "walking" contain multiple actors in close proximity, which results in false detections due to occlusions. Additionally, walking is a very slow action with fine grained features and potentially lacks enough temporal displacement in 8 frames to be picked up by our detector due to sparse temporal sampling strategy. Ultimately, our approach is off by only 2\% when compared to the State-of-the-Art approaches that utilize either multi-modal, 3-dimensional, or complex proposal architecture solutions. The State-of-the-Art comparison in terms of mean Average precision (mAP) is summarised in Table \ref{tab:Table2}.
\setlength{\tabcolsep}{4pt} 
\renewcommand{\arraystretch}{1.5}
\begin{table}[ht]
\centering
\large
\caption{State-of-the-Art per class frame mAP comparison in UCF-Sports}\label{tab:Table1}
\resizebox{\columnwidth}{!}{\begin{tabular}{c|c|c|c|c|c}
\Xhline{2\arrayrulewidth}
\textbf{Action Class} &
\textbf{[\citeauthor{action_tubes}]} & 
\textbf{[\citeauthor{learning}]} & 
\textbf{[\citeauthor{Multiregion}]} & 
\textbf{[\citeauthor{TCNN}]} & 
\textbf{S-RAD} \\
\Xhline{2\arrayrulewidth}

Diving &75.79 &60.71 &96.12 &84.37 &\textbf{99.90} \\

\cellcolor[HTML]{F5F5F5}Golf &\cellcolor[HTML]{F5F5F5}69.29 &\cellcolor[HTML]{F5F5F5}77.54 &\cellcolor[HTML]{F5F5F5}80.46 &\cellcolor[HTML]{F5F5F5}90.79 &\cellcolor[HTML]{F5F5F5}87.20 \\

Kicking &54.60 &65.26 &73.48 &86.48 &76.00 \\
\cellcolor[HTML]{F5F5F5}Lifting &\cellcolor[HTML]{F5F5F5}99.09 &\cellcolor[HTML]{F5F5F5}100.00 &\cellcolor[HTML]{F5F5F5}99.17 &\cellcolor[HTML]{F5F5F5}99.76 &\cellcolor[HTML]{F5F5F5}\textbf{99.96} \\
Riding &89.59 &99.53 &97.56 &100.0 &\textbf{99.90} \\
\cellcolor[HTML]{F5F5F5}Run &\cellcolor[HTML]{F5F5F5}54.89 &\cellcolor[HTML]{F5F5F5}52.60 &\cellcolor[HTML]{F5F5F5}82.37 &\cellcolor[HTML]{F5F5F5}83.65 &\cellcolor[HTML]{F5F5F5}\textbf{89.79} \\
Skate Boarding &29.80 &47.14 &57.43 &68.71 &67.93 \\
\cellcolor[HTML]{F5F5F5}Swing1 &\cellcolor[HTML]{F5F5F5}88.70 &\cellcolor[HTML]{F5F5F5}88.87 &\cellcolor[HTML]{F5F5F5}83.64 &\cellcolor[HTML]{F5F5F5}65.75 &\cellcolor[HTML]{F5F5F5}\textbf{88.78} \\
Swing2 &74.50 &62.85 &98.50 &99.71 &\textbf{99.9} \\
\cellcolor[HTML]{F5F5F5}Walk &\cellcolor[HTML]{F5F5F5}44.70 &\cellcolor[HTML]{F5F5F5}64.43 &\cellcolor[HTML]{F5F5F5}75.98 &\cellcolor[HTML]{F5F5F5}87.79 &\cellcolor[HTML]{F5F5F5}40.71 \\
\Xhline{2\arrayrulewidth}

\end{tabular}}
\end{table}

\begin{figure}[!h]
    \centering
     \includegraphics[width=1\linewidth]{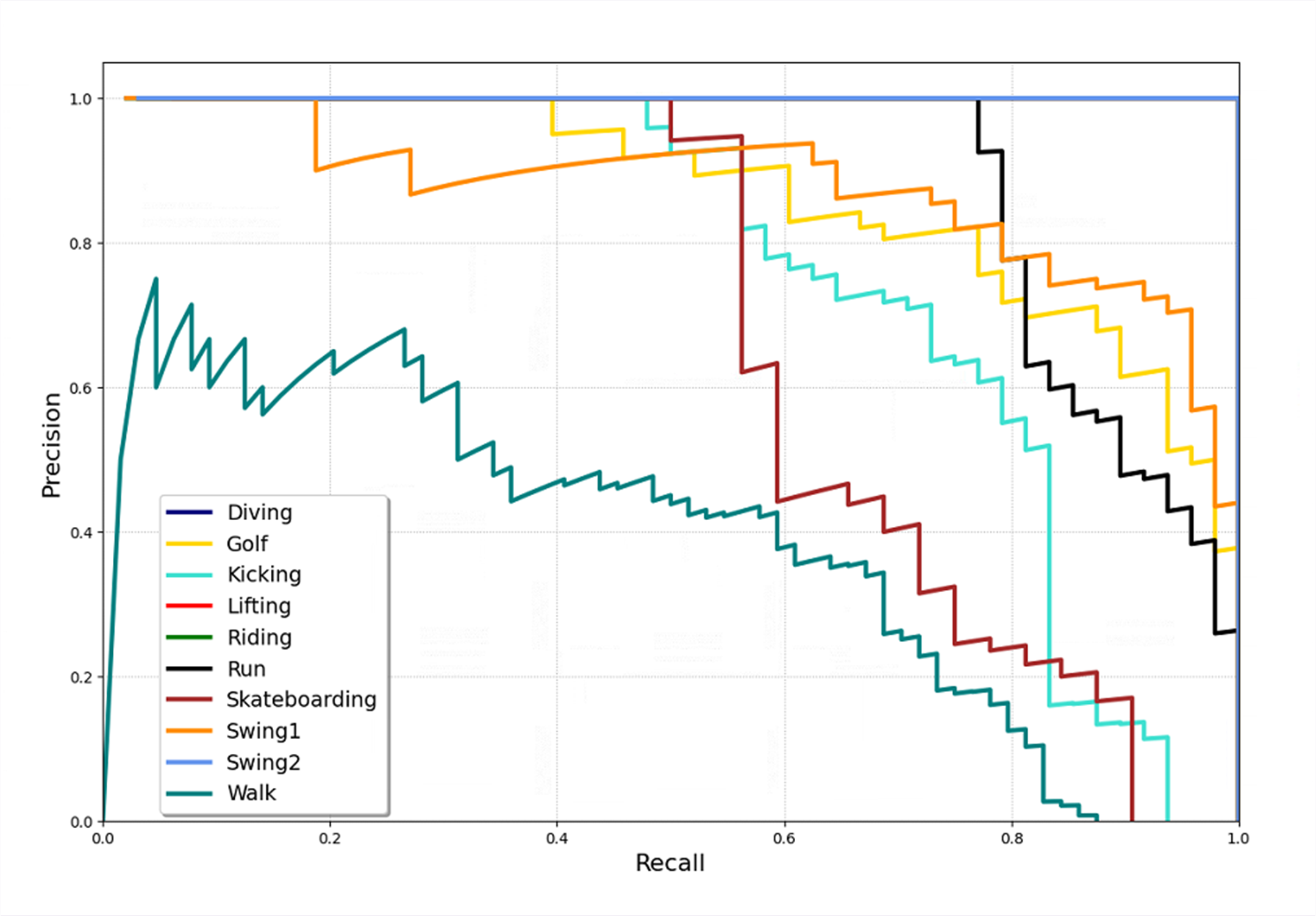}
    \caption{Precision-Recall curve per Action class in UCF-Sports}
    \label{fig:PRcurve}
\end{figure}

\begin{table}[http]
\LARGE
\setlength{\tabcolsep}{8pt} 
\renewcommand{\arraystretch}{1.2}
\centering
\caption{Overall frame mAP at IOU 0.5 threshold comparison in UCF-Sports Action dataset}\label{tab:Table2}
\begin{adjustbox}{width=1\linewidth,center}
{\begin{tabular}{c|c|c|c|c|c|c|c}
\textbf{ } &
{\textbf{[\citeyear{action_tubes}]}} & 
{\textbf{[\citeyear{learning}]}} & 
{\textbf{[\citeyear{Multiregion}]}} & 
{\textbf{[\citeyear{TCNN}]}} & 
{\textbf{[\citeyear{ACTdetector}]}}& 
{\textbf{[\citeyear{videocapsule_net}]}}&
{\textbf{{S-RAD}}} \\
\Xhline{2\arrayrulewidth}
\hline
{\textbf{mAP}} &{68.09} &{71.90} &{84.51} &{86.70} &{\textbf{87.7}} &{83.9} &{85.04} \\

\end{tabular}}
\end{adjustbox}
\end{table}
The Precision Recall AUC is ploted in Figure \ref{fig:PRcurve} shows the capability of our algorithm to separate different classes.

\begin{figure}[!h]
    \centering
     \includegraphics[width=1\linewidth]{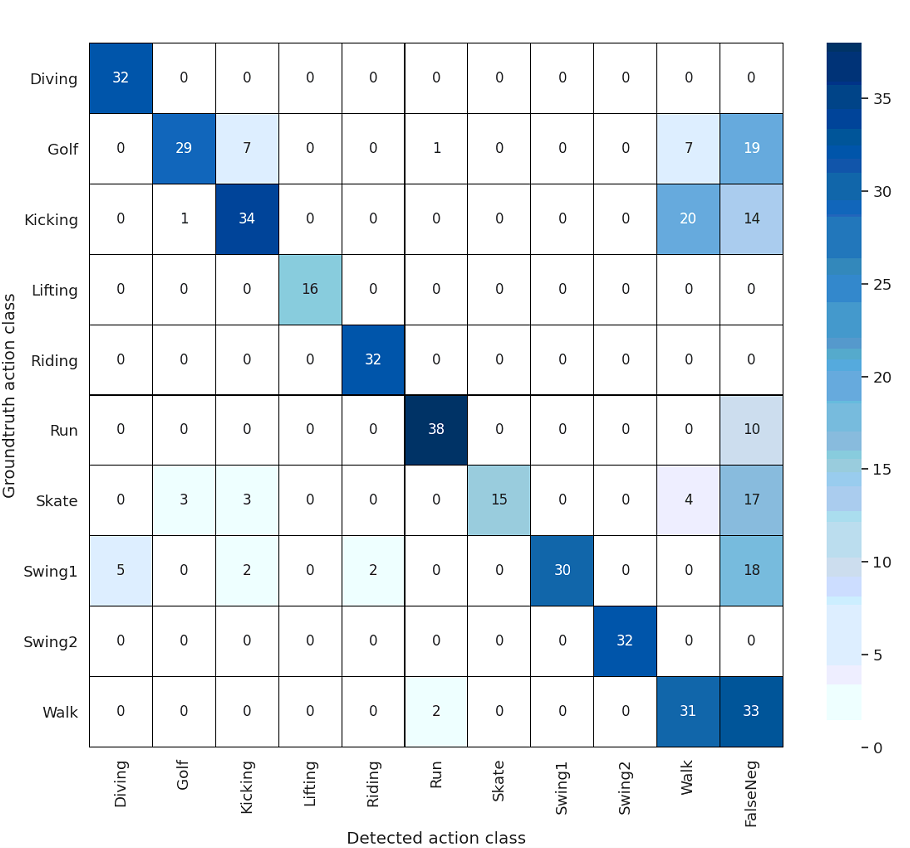}
    \caption{Confusion matrix of S-RAD on UCF-Sports}
    \label{fig:confmatrixucfsport}
\end{figure}

We also provided the confusion matrix  to better understand the detections with the original ground truth in Figure \ref{fig:confmatrixucfsport}. 
The confusion matrix is calculated considering both the detection and classification tasks. Here the grids in the diagonal are the true positive's whose IOU$>$0.5 and the detected action class label match with the ground truth action class label. Other columns are the false positive whose IOU$>$0.5 but the detected action class label does not match the ground truth action class label. The last column contains false negatives with detections with an IOU$<$0.5.

\subsection{Results on UR Fall Dataset}
We have also evaluated our framework on the healthcare extensive dataset \cite{urfall}. The UR Fall dataset is composed of 70 videos: (i) 30 videos of falls; and (ii) 40 videos displaying diverse activities. We used \cite{mmdetection} pre-trained only on the person class in the coco dataset to obtain the bounding box annotations for the ground truth. On average there are 56 videos in the training and 14 videos are in the testing dataset.

For the UR Fall dataset we calculate specificity, sensitivity and accuracy along with mAP for comparison.

(1)Sensitivity: A metric to evaluate detecting falls. And compute the ratio of trues positives to the number of falls.

\begin{equation}
\small
Sensitivity=\frac{TP}{TP+FN}*100
\end{equation}
(2)Specificity: A metric to evaluate how much our algorithm detects
just "fall" and avoids misclassification with the "not fall" class.
\begin{equation}
\small
Specificity=\frac{TN}{TN+FP}*100
\end{equation}
(3)Accuracy: Metric to compute how much our algorithm can differ
between falls and non-fall videos.
\begin{equation}
\small
Accuracy=\frac{TP+TN}{TN+FP+TP+FN}*100
\end{equation}

 True positive (TP) means that the frame has a fall and our algorithm has detected fall in those frames.True negative (TN) refers to the frames that don’t contain fall and our algorithm does not detect fall in those frames. False negative (FN) designates the frames containing falls, however our algorithm fails to detect the fall in those frames. Finally, false positive (FP) indicates the frames don’t contain a fall, yet our algorithm claims to detect a fall. For the sake of comparison with the other classification based State-of-the-Art papers we take the detection with the highest confidence score from the output of S-RAD and compare it's class label with the ground truth class label to calculate the above mentioned parameters. Since our approach is based on frame level detection, the classification task on UR fall dataset is also done in frame level. We achieved a competitive score of 96.54 \% in mAP (detection task at frame level). It is important to note, other State-of-the-Art approaches on this dataset relied solely on classification, hence our comparison being concentrated on the classification metrics. The Results are shown on Table \ref{tab:Table3}, showing S-RAD's true capabilities in the field of healthcare.
 
\begin{figure}[!h]
    \centering
     \includegraphics[width=0.8\linewidth]{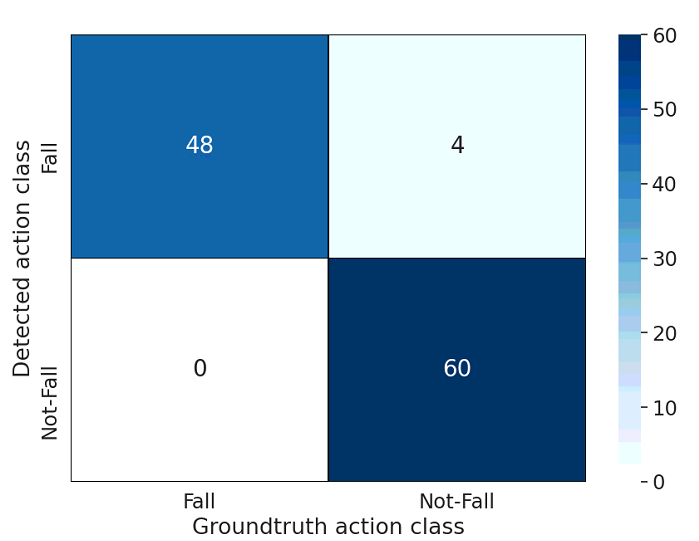}
    \caption{Confusion matrix of S-RAD on UR Fall dataset }
    \label{fig:URconfusionmatrix}
\end{figure}

The confusion matrix on Figure \ref{fig:URconfusionmatrix} shows the ability of the S-RAD to distinguish Fall and Not Fall with only 4 instances being misclassified as Fall. 

\begin{table}[ht]
\setlength{\tabcolsep}{3pt} 
\renewcommand{\arraystretch}{1.5}
\centering
\large
\caption{State-of-the-Art per frame comparison in UR Fall dataset}\label{tab:Table3}
\resizebox{\columnwidth}{!}{\begin{tabular}{c|c|c|c|c|c}
\Xhline{2\arrayrulewidth}
\textbf{} &
\textbf{[\citeauthor{Alaouikeypointfalldetection}]}& 
\textbf{[\citeauthor{3Dfalldetection}]} & 
\textbf{[\citeauthor{Cameiro}]} &
\textbf{[\citeauthor{Threestreamfall}]} &
\textbf{S-RAD}\\
\Xhline{2\arrayrulewidth}

Sensitivity &100 &- &100 &100 &\textbf{100} \\
\cellcolor[HTML]{F5F5F5}Specificity &\cellcolor[HTML]{F5F5F5}95 &\cellcolor[HTML]{F5F5F5}-
&\cellcolor[HTML]{F5F5F5}98.61 &\cellcolor[HTML]{F5F5F5}98.77 &\cellcolor[HTML]{F5F5F5}\textbf{93.75}\\

Accuracy &97.5 &99.27 &98.77 &98.84 &\textbf{96.46}\\
\Xhline{2\arrayrulewidth}
\end{tabular}}
\end{table}
  
\subsection{Real-time Execution}

The S-RAD framework has the advantage of reduced inference time and less number of parameters, enabling us to perform real-time on the edge activity monitoring in a privacy-aware manner. We compare our framework with others in terms of FPS (Frame-Per-Second) and mAP in Table {\ref{tab:Table3 }} on the UCF-Sports Action dataset. We tested our models on one Titan V GPU (except the work of TubeCNN \cite{TCNN}, which was reported on a titan X). The trade-off is between accuracy and inference FPS, as well as parameters. Among the state of the art approaches, our method has the second fastest run time and can process ~41 frames per second which is three times faster than \cite{TCNN} and \cite{Multiregion}. Moreover, the number of parameters of our framework is the smallest, about 28.36 M in Table \ref{tab:Table3 }, although works like \cite{videocapsule_net} have better FPS with their models, their features are too heavy to fit into a real-time edge device, additionally our work maintains a higher mAP at a high resolution when compared to their work. We were unable to provide performance comparisons with the State-of-the-Art approaches on the UR Fall dataset as most of the approaches are not publicly available to run on the edge device, and do not provide performance metrics of their own.

\begin{table}[!h]
\LARGE
\setlength{\tabcolsep}{12pt} 
\renewcommand{\arraystretch}{3}
\caption{Comparison on Server Class Execution on Nvidia Titan platform}\label{tab:Table3 }
\begin{adjustbox}{width=1.0\linewidth,center}
{\begin{tabular}{c|c|c|c|c|c}
\Xhline{2\arrayrulewidth}
{\fontsize{22}{7.2}\selectfont\textbf{Approach}} &
{\fontsize{22}{7.2}\selectfont\textbf{Input}} &

{\fontsize{22}{7.2}\selectfont\textbf{Resolution}} &
{\fontsize{22}{7.2}\selectfont\textbf{Param \# (M)}} &
{\fontsize{22}{7.2}\selectfont\textbf{FPS}} &
{\fontsize{22}{7.2}\selectfont\textbf{mAP}} \\
\Xhline{2\arrayrulewidth}
{\fontsize{20}{7.2}\selectfont\cellcolor[HTML]{F5F5F5}Multi-stream \cite{Multiregion}} &{\fontsize{20}{7.2}\selectfont\cellcolor[HTML]{F5F5F5}RGB+Flow}

&{\fontsize{20}{7.2}\selectfont\cellcolor[HTML]{F5F5F5}600x1067} &
{\fontsize{20}{7.2}\selectfont\cellcolor[HTML]{F5F5F5}274} &
{\fontsize{20}{7.2}\selectfont\cellcolor[HTML]{F5F5F5}~11.82} &
{\fontsize{20}{7.2}\selectfont\cellcolor[HTML]{F5F5F5}84.51} \\
{\fontsize{20}{7.2}\selectfont CapsuleNet\cite{videocapsule_net}} &{\fontsize{20}{7.2}\selectfont RGB} &

{\fontsize{20}{7.2}\selectfont112x112} &
{\fontsize{20}{7.2}\selectfont103.137} &
{\fontsize{20}{7.2}\selectfont78.41} &
{\fontsize{20}{7.2}\selectfont83.9} \\

{\fontsize{20}{7.2}\selectfont\cellcolor[HTML]{F5F5F5}TubeCNN\cite{TCNN}} &
{\fontsize{20}{7.2}\selectfont\cellcolor[HTML]{F5F5F5}RGB} &

{\fontsize{20}{7.2}\selectfont\cellcolor[HTML]{F5F5F5}300x400} &
{\fontsize{20}{7.2}\selectfont\cellcolor[HTML]{F5F5F5}245.87} &
{\fontsize{20}{7.2}\selectfont\cellcolor[HTML]{F5F5F5}~17.391} &
{\fontsize{20}{7.2}\selectfont\cellcolor[HTML]{F5F5F5}86.7} \\

{\fontsize{20}{7.2}\selectfont ACT\cite{ACTdetector}} &
{\fontsize{20}{7.2}\selectfont RGB+Flow} &
{\fontsize{20}{7.2}\selectfont 300x300} &
{\fontsize{20}{7.2}\selectfont 50} &
{\fontsize{20}{7.2}\selectfont~12} &
{\fontsize{20}{7.2}\selectfont 87.7} \\

\Xhline{2\arrayrulewidth}
\textbf{{\fontsize{20}{7.2}\selectfont\cellcolor[HTML]{F5F5F5}S-RAD}} &
\textbf{{\fontsize{20}{7.2}\selectfont\cellcolor[HTML]{F5F5F5}RGB}} &
\textbf{{\fontsize{20}{7.2}\selectfont\cellcolor[HTML]{F5F5F5}300x400}} &
\textbf{{\fontsize{20}{7.2}\selectfont\cellcolor[HTML]{F5F5F5}28.35}} &
\textbf{{\fontsize{20}{7.2}\selectfont\cellcolor[HTML]{F5F5F5}41.64}} &
\textbf{{\fontsize{20}{7.2}\selectfont\cellcolor[HTML]{F5F5F5}85.04}} \\

\bottomrule

\end{tabular}}
\end{adjustbox}
\end{table}

We additionally evaluated our work on an edge platform, the Nvidia Xavier to test its performance on an resource constrained edge platform. We compare the work of VideoCapsuleNet \cite{videocapsule_net} with our approach, and despite their initial performance advantage on the Titan V, our work is the only model capable of running on the memory constrained edge device. S-RAD, as opposed to VideoCapsuleNet folds temporal data into the channel dimension, and as a result avoids introducing another dimension to the tensor sizes. VideoCapsuleNet not only process 3D spatial-temporal feature maps, but they also introduce another dimension of complexity in the form of capsules. We also observed 6.0 FPS with only 5.21W of total SoC (on chip) power consumption.

\section{Conclusion}
This paper introduced a novel Single Run Action detector (S-RAD) for activity monitoring. S-RAD provides end-to-end action detection without the use of computationally heavy methods with the ability for real-time execution of embedded edge devices. S-RAD is a privacy-preserving approach and inherently protects Personally Identifiable Information (PII). Results on UCF-Sports and UR Fall dataset presented comparable accuracy to State-of-the-Art approaches with significantly lower model size and computation demand and the ability for real-time execution on edge embedded device.

\bibliographystyle{named}
\bibliography{main.bib}

\begin{thebibliography}{}

\bibitem[\protect\citeauthoryear{{Alaoui} \bgroup \em et al.\egroup
  }{2019}]{Alaouikeypointfalldetection}
A.~Y. {Alaoui}, S.~{El Fkihi}, and R.~O.~H. {Thami}.
\newblock Fall detection for elderly people using the variation of key points
  of human skeleton.
\newblock {\em IEEE Access}, 7:154786--154795, 2019.

\bibitem[\protect\citeauthoryear{Atallah \bgroup \em et al.\egroup
  }{2011}]{atallah2011sensor}
Louis Atallah, Benny Lo, Rachel King, and Guang-Zhong Yang.
\newblock Sensor positioning for activity recognition using wearable
  accelerometers.
\newblock {\em IEEE transactions on biomedical circuits and systems},
  5(4):320--329, 2011.

\bibitem[\protect\citeauthoryear{{Cameiro} \bgroup \em et al.\egroup
  }{2019}]{Cameiro}
S.~A. {Cameiro}, G.~P. {da Silva}, G.~V. {Leite}, R.~{Moreno}, S.~J.~F.
  {Guimarães}, and H.~{Pedrini}.
\newblock Multi-stream deep convolutional network using high-level features
  applied to fall detection in video sequences.
\newblock In {\em 2019 International Conference on Systems, Signals and Image
  Processing (IWSSIP)}, pages 293--298, 2019.

\bibitem[\protect\citeauthoryear{Chen \bgroup \em et al.\egroup
  }{2019}]{mmdetection}
Kai Chen, Jiaqi Wang, Jiangmiao Pang, Yuhang Cao, Yu~Xiong, Xiaoxiao Li,
  Shuyang Sun, Wansen Feng, Ziwei Liu, Jiarui Xu, Zheng Zhang, Dazhi Cheng,
  Chenchen Zhu, Tianheng Cheng, Qijie Zhao, Buyu Li, Xin Lu, Rui Zhu, Yue Wu,
  Jifeng Dai, Jingdong Wang, Jianping Shi, Wanli Ouyang, Chen~Change Loy, and
  Dahua Lin.
\newblock {MMDetection}: Open mmlab detection toolbox and benchmark.
\newblock {\em arXiv preprint arXiv:1906.07155}, 2019.

\bibitem[\protect\citeauthoryear{Duarte \bgroup \em et al.\egroup
  }{2018}]{videocapsule_net}
Kevin Duarte, Yogesh~S Rawat, and Mubarak Shah.
\newblock Videocapsulenet: A simplified network for action detection.
\newblock In {\em Proceedings of the 32nd International Conference on Neural
  Information Processing Systems}, NIPS’18, page 7621–7630, Red Hook, NY,
  USA, 2018. Curran Associates Inc.

\bibitem[\protect\citeauthoryear{Everingham \bgroup \em et al.\egroup
  }{2010}]{pascal_voc}
Mark Everingham, Luc Gool, Christopher~K. Williams, John Winn, and Andrew
  Zisserman.
\newblock The pascal visual object classes (voc) challenge.
\newblock {\em Int. J. Comput. Vision}, 88(2):303–338, June 2010.

\bibitem[\protect\citeauthoryear{Gkioxari and Malik}{2015}]{action_tubes}
Georgia Gkioxari and Jitendra Malik.
\newblock Finding action tubes.
\newblock In {\em Proceedings of the IEEE conference on computer vision and
  pattern recognition}, pages 759--768, 2015.

\bibitem[\protect\citeauthoryear{He \bgroup \em et al.\egroup }{2015}]{resnet}
Kaiming He, Xiangyu Zhang, Shaoqing Ren, and Jian Sun.
\newblock Deep residual learning for image recognition.
\newblock {\em CoRR}, abs/1512.03385, 2015.

\bibitem[\protect\citeauthoryear{Hou \bgroup \em et al.\egroup }{2017}]{TCNN}
Rui Hou, Chen Chen, and Mubarak Shah.
\newblock Tube convolutional neural network (t-cnn) for action detection in
  videos.
\newblock In {\em The IEEE International Conference on Computer Vision (ICCV)},
  Oct 2017.

\bibitem[\protect\citeauthoryear{Kalogeiton \bgroup \em et al.\egroup
  }{2017}]{ACTdetector}
Vicky Kalogeiton, Philippe Weinzaepfel, Vittorio Ferrari, and Cordelia Schmid.
\newblock Action tubelet detector for spatio-temporal action localization.
\newblock {\em CoRR}, abs/1705.01861, 2017.

\bibitem[\protect\citeauthoryear{Kay \bgroup \em et al.\egroup
  }{2017}]{kay2017kinetics}
Will Kay, Joao Carreira, Karen Simonyan, Brian Zhang, Chloe Hillier, Sudheendra
  Vijayanarasimhan, Fabio Viola, Tim Green, Trevor Back, Paul Natsev, et~al.
\newblock The kinetics human action video dataset.
\newblock {\em arXiv preprint arXiv:1705.06950}, 2017.

\bibitem[\protect\citeauthoryear{Kwolek and Kepski}{2014}]{urfall}
B.~Kwolek and Michal Kepski.
\newblock Human fall detection on embedded platform using depth maps and
  wireless accelerometer.
\newblock {\em Computer methods and programs in biomedicine}, 117 3:489--501,
  2014.

\bibitem[\protect\citeauthoryear{{Leite} \bgroup \em et al.\egroup
  }{2019}]{Threestreamfall}
G.~{Leite}, G.~{Silva}, and H.~{Pedrini}.
\newblock Fall detection in video sequences based on a three-stream
  convolutional neural network.
\newblock In {\em 2019 18th IEEE International Conference On Machine Learning
  And Applications (ICMLA)}, pages 191--195, 2019.

\bibitem[\protect\citeauthoryear{Lin \bgroup \em et al.\egroup }{2018}]{tsm}
Ji~Lin, Chuang Gan, and Song Han.
\newblock Temporal shift module for efficient video understanding.
\newblock {\em CoRR}, abs/1811.08383, 2018.

\bibitem[\protect\citeauthoryear{Liu \bgroup \em et al.\egroup }{2015}]{SSD}
Wei Liu, Dragomir Anguelov, Dumitru Erhan, Christian Szegedy, Scott~E. Reed,
  Cheng{-}Yang Fu, and Alexander~C. Berg.
\newblock {SSD:} single shot multibox detector.
\newblock {\em CoRR}, abs/1512.02325, 2015.

\bibitem[\protect\citeauthoryear{{Lu} \bgroup \em et al.\egroup
  }{2019}]{3Dfalldetection}
N.~{Lu}, Y.~{Wu}, L.~{Feng}, and J.~{Song}.
\newblock Deep learning for fall detection: Three-dimensional cnn combined with
  lstm on video kinematic data.
\newblock {\em IEEE Journal of Biomedical and Health Informatics},
  23(1):314--323, 2019.

\bibitem[\protect\citeauthoryear{Mirzadeh and
  Ghasemzadeh}{2020}]{Mirzadeh2020OptDeploy}
Seyed~Iman Mirzadeh and Hassan Ghasemzadeh.
\newblock Optimal policy for deployment of machine learning modelson
  energy-bounded systems.
\newblock In {\em Proceedings of the Twenty-Ninth International Joint
  Conference on Artificial Intelligence (IJCAI)}, 2020.

\bibitem[\protect\citeauthoryear{{Neff} \bgroup \em et al.\egroup
  }{2020}]{REVAMP2T}
C.~{Neff}, M.~{Mendieta}, S.~{Mohan}, M.~{Baharani}, S.~{Rogers}, and
  H.~{Tabkhi}.
\newblock Revamp2t: Real-time edge video analytics for multicamera
  privacy-aware pedestrian tracking.
\newblock {\em IEEE Internet of Things Journal}, 7(4):2591--2602, 2020.

\bibitem[\protect\citeauthoryear{Pagan \bgroup \em et al.\egroup
  }{2018}]{pagan2018toward}
Josue Pagan, Ramin Fallahzadeh, Mahdi Pedram, Jose~L Risco-Martin, Jose~M Moya,
  Jose~L Ayala, and Hassan Ghasemzadeh.
\newblock Toward ultra-low-power remote health monitoring: An optimal and
  adaptive compressed sensing framework for activity recognition.
\newblock {\em IEEE Transactions on Mobile Computing (TMC)}, 18(3):658--673,
  2018.

\bibitem[\protect\citeauthoryear{Peng and Schmid}{2016}]{Multiregion}
Xiaojiang Peng and Cordelia Schmid.
\newblock Multi-region two-stream r-cnn for action detection.
\newblock In {\em European conference on computer vision}, pages 744--759.
  Springer, 2016.

\bibitem[\protect\citeauthoryear{Ren \bgroup \em et al.\egroup
  }{2015}]{Faster-RCNN}
Shaoqing Ren, Kaiming He, Ross~B. Girshick, and Jian Sun.
\newblock Faster {R-CNN:} towards real-time object detection with region
  proposal networks.
\newblock {\em CoRR}, abs/1506.01497, 2015.

\bibitem[\protect\citeauthoryear{Saeedi \bgroup \em et al.\egroup
  }{2014}]{saeedi2014toward}
Ramyar Saeedi, Janet Purath, Krishna Venkatasubramanian, and Hassan
  Ghasemzadeh.
\newblock Toward seamless wearable sensing: Automatic on-body sensor
  localization for physical activity monitoring.
\newblock In {\em 2014 36th Annual International Conference of the IEEE
  Engineering in Medicine and Biology Society}, pages 5385--5388. IEEE, 2014.

\bibitem[\protect\citeauthoryear{Simonyan and Zisserman}{2014}]{Twostream}
Karen Simonyan and Andrew Zisserman.
\newblock Two-stream convolutional networks for action recognition in videos.
\newblock {\em CoRR}, abs/1406.2199, 2014.

\bibitem[\protect\citeauthoryear{Soomro and Zamir}{2014}]{UCFsports}
Khurram Soomro and Amir~Roshan Zamir.
\newblock Action recognition in realistic sports videos.
\newblock 2014.

\bibitem[\protect\citeauthoryear{Tran \bgroup \em et al.\egroup }{2014}]{c3D}
Du~Tran, Lubomir~D. Bourdev, Rob Fergus, Lorenzo Torresani, and Manohar Paluri.
\newblock {C3D:} generic features for video analysis.
\newblock {\em CoRR}, abs/1412.0767, 2014.

\bibitem[\protect\citeauthoryear{Wang \bgroup \em et al.\egroup }{2016}]{TSN}
Limin Wang, Yuanjun Xiong, Zhe Wang, Yu~Qiao, Dahua Lin, Xiaoou Tang, and
  Luc~Van Gool.
\newblock Temporal segment networks: Towards good practices for deep action
  recognition.
\newblock {\em CoRR}, abs/1608.00859, 2016.

\bibitem[\protect\citeauthoryear{Weinzaepfel \bgroup \em et al.\egroup
  }{2015}]{learning}
Philippe Weinzaepfel, Za{\"{\i}}d Harchaoui, and Cordelia Schmid.
\newblock Learning to track for spatio-temporal action localization.
\newblock {\em CoRR}, abs/1506.01929, 2015.

\end{thebibliography}

\end{document}